\documentclass{article}
\usepackage{amssymb,amsmath,amsthm}
\usepackage{graphicx}
\usepackage{indentfirst} 
\usepackage{cmap}

\usepackage[utf8]{inputenc}
\usepackage[russian]{babel}

\usepackage{geometry}
\usepackage{eulervm}
\usepackage{graphicx}
\usepackage{multirow}

\usepackage[nosort]{cite}
\usepackage{color}

\usepackage{mathtools}

\DeclarePairedDelimiter\floor{\lfloor}{\rfloor}

\usepackage{wrapfig}
\graphicspath{{Pics/}}

\usepackage{amsthm}
\usepackage{amsfonts}

\newtheorem{mydef}{Определение}
\newtheorem{mystate}{Утверждение}
\newtheorem{mystatet}{Утверждение}
\newtheorem{mylemm}{Лемма}
\newtheorem{mytheo}{Теорема}

\newcommand{\lal}{\scalebox{3}[1]{$\propto$}}
\newcommand{\revlal} {$\overline{\reflectbox{\lal}}$}
\newcommand{\rotlal}{\rotatebox[origin=c]{-90}{\lal}}

\newcommand{\vave}[1]{$\overset{#1}{\scalebox{3}[1]{$\sim$}}$}
\newcommand{\vavep}[3]{$\overset{#1}{^#2\scalebox{3}[1]{$\sim$}^#3}$}

\newcommand{\randcore}{\scalebox{3}[1]{\rotatebox[origin=c]{-90}{$\dagger$}}}

\newcommand{\rands}{\randcore}
\newcommand{\revrands} {$\overline{\reflectbox{\randcore}}$}

\DeclareMathOperator*{\argmax}{argmax}

\begin{document}

\title{Исследование свойств диадического паттерна быстрого преобразования Хафа}

\author{Е.И. Ершов \\ ИППИ РАН, МФТИ (ГУ) \\ ershov@iitp.ru \and С.М. Карпенко \\ ИППИ РАН \\ simon.karpenko@gmail.com}

\maketitle

\section*{Аннотация}

Одним из востребованных алгоритмов в области обработки изображений является быстрое преобразование Хафа (БПХ).
Эффективность БПХ во многом связана с использованием дискретных прямых специального вида - диадических паттернов. 
В работе исследуются свойства структуры данного паттерна, приводится ряд экспериментальных исследований его свойств. 
Доказана теорема об асимптотике пиковой ошибки аппроксимации диадическим паттерном соответствующей прямой линии: $O(log(n)/6)$, где  $n$ -- линейный размер изображения.

\section{Введение}

Преобразование Хафа (ПХ) изобретено Полом Хафом для анализа фотографий, полученных в пузырьковой камере, в 1959 году. 
Запатентовано же оно было автором в 1962 году \cite{Hough1962patent}.
Как инструмент анализа изображений ПХ получило широкое распрастранение \cite{illingworth1988survey}.
Важнейшим частным случаем ПХ является преобразование Хафа для прямых; ниже рассматривается только этот случай.
Для квадратного изображения со стороной $n$ вычислительная сложность преобразования составляет $O(n^3)$ операций (на изображении $O(n^2)$ прямых, длина каждой $O(n)$). 
Позднее появился более эффективный алгоритм вычисления аппроксиации преобразования Хафа, так называемое быстрое преобразование Хафа с вычислительной сложностью $O(n^2 \log n)$ операций.

История изобретений и переизобретений алгоритма БПХ заслуживает отдельного внимания. 
Впервые алгормтм был предложен австрийским ученым Готсом в его диссертации 1993 года \cite{gots1993dis}, позднее, в 1995, работа была опубликована на английском языке \cite{Gotz1995}.
Вторым изобретателем стал французский профессор информатики Вулемин в 1994 \cite{vuillemin1994} году. 
Позднее, алгоритм был переоткрыт в 1998 \cite{Brady1998}, в 2004 \cite{NiksKarp2004} и в 2005 \cite{frederick2005real} годах.
Судя по тому, что в современном обзоре алгоритмов вычисления различных преобразований Хафа по-прежнему не фигурирует быстрая версия для прямых \cite{mukhopadhyay2015survey}, можно ожидать новых переоткрытий.

С другой стороны такое количество переоткрытий данного инструмента свидетельствует также и о его востребованности.
Можно указать множество применений ПХ, среди которых можно выделить: детектирование прямолинейных краев, определение ориентации документа, определение точек схода линий \cite{NiksKarp2008} и др. 
Помимо прочего, преобразование Хафа успешно применяется в робастном регрессионном анализе  \cite{Goldenshluger, Ballester1994, bezm2012rus}.

Математически преобразование Хафа - это дискретная форма преобразования Радона.
С точки зрения обработки изображений, прямая является набором пикселей, апроксимирующих соответствующую прямую линию в области изображения.
Такой набор пикселей будем далее называть дискретной прямой или паттерном.
Важно отметить, что точность аппроксимации дискретной прямой непосредственно зависит от метода дискретизации и от способа её построения. 
Так, при вычислении БПХ прямые аппроксимируются паттернами специального вида - \textit{диадическими}. 
Способ построения диадического паттерна представлен, например, в \cite{NiksKarp2008} и будет изложен ниже. 

Теоретическое же исследование точностных характеристик БПХ \cite{gots1993dis} является открытым вопросом.
Далее будут предложены два определения диадического паттерна и показана их эквивалентность.
С использованием нового определения доказана теорема об асимптотике пиковой ошибки аппроксимации диадическим паттерном.

\section{Параметризация, структура и ошибка аппроксимации диадического паттерна}

Пусть ноль системы координат изображения помещен в центр левого верхнего пикселя, вертикальная ось $y$ направлена вниз, горизонтальная ось $x$ и направлена вправо. 
При описании структуры диадического паттерна принято разделать их на 4 типа по значению $k$ в уравнении прямой с уловым коэффициентов $y = kx + b$: преимущественно-горизонтальные с наклоном влево ($k \in [-1; 0]$),
преимущественно-горизонтальные с наклоном вправо ($k \in [0; 1]$), преимущественно-вертикальные с наклоном вправо ($k \in [1; \infty]$), преимущественно-вертикальные с наклоном влево ($\alpha \in [-\infty; -1]$). 
Вычисление БПХ для различных типов паттернов производится независимо, результирующее пространство Хафа для всего изображения получается путем сшивки результатов вычисления БПХ для каждого типа прямых. 
В силу симметрии, для анализа точности аппроксимации диадическим паттерном достаточно рассматривать только один тип, например, будем рассматривать преимущественно-горизонтальные прямые с наклоном вправо.
Для таких паттернов существует две точки на границах изображения вида $P1(0, s), P2(n-1, t+s)$, где $s$  параметризует смещение(сдвиг) прямой относительно начала координат, а $t$ - наклон прямой, измеренный в пикселях.

Кратко изложим способ построения диадического паттерна.
В работе рассмотрим квадратное изображения размера $n = 2^p$, где $p \in \mathbb{Z}$.

Для построения диадического паттерна разделим изображение на две равные части вертикальной прямой.
Для изображения с линейным размером $2^p$ пикселей такая линия всегда будет совпадать с границами пикселей в области изображения.
Выберем два пикселя, сверху и снизу от разделяющей прямой с координатами  $(2^{p-1} - 1, \floor{t/2})$ и $(2^{p-1}, t \mod 2 + \floor{t/2})$ соответственно.
Далее процедура повторяется рекурсивно для левого и правого отрезка относительно разделяющих эти отрезки вертикальных линий до тех пор пока не выполнится условие $t = 1$.
Заметим, что величины наклонов на двух последовательных шагах рекурсии связаны следующим образом:

\begin{equation}
    t_k = \floor[\bigg]{\frac{t_{k-1}}{2}} +  \bigg(t_{k-1}\mod 2\bigg) + \floor[\bigg]{\frac{t_{k-1}}{2}},
    \label{eq:rec_def}
\end{equation}
где $k$ - шаг рекурсии.

\begin{mydef}
Дискретная прямая, построенная согласно изложенному выше способу, называется \textbf{диадическим паттерном}. 
\end{mydef}

Заметим, что структура диадической прямой не зависит от сдвига $s$, а определяется только наклоном $t$.
Будем рассматривать далее только паттерны с $s = 0$.

Вышеизложенная процедура и её вариации \cite{Gotz1995, vuillemin1994} неудобны для анализа ошибки аппроксимации в аналитическом виде.
Дадим альтернативное определение \cite{ershov2015dev} с использованием базисных диадических прямых.

\begin{mydef}
\textit{Базисной диадической прямой} называется дискретная прямая вида

\begin{equation*}
    y = D_r(x) = \bigg[\frac{t x}{2^p - 1}\bigg] = \bigg[\frac{2^r x}{2^p - 1}\bigg],
    \label{eq:anal_def}
\end{equation*}
где $x, y, t, r \in \mathbb{Z}^+$ абсцисса, ордината и параметр величины наклона соответственно, причем $x, y, t < n$, а $r < p$ .
\end{mydef}

\begin{mystate}
Для $t = 2^r$ множество пикселей, полученных согласно определению 1 и 2 совпадают.
\end{mystate}
\begin{proof}
Проанализируем определение диадического паттерна для базисной прямой по правилу \ref{eq:rec_def}. 

Очевидно, что $t \mod 2 = 1$ при $t = 2^r$ только на $r$-м шаге рекурсии для отрезка длины $(2^p - 1)/2^r$.
Множество абсцисс разделяющих линий для каждого такого отрезка выражается формулой $(m + \frac{1}{2}) \frac{2^p - 1}{2^r}$, где $m \in [0, 2^r - 1]$.
Тогда определим множество абсцисс пикселей, располагающихся по одну сторону от разделяющих прямых, как $\hat{X} = \floor{(m + \frac{1}{2}) \frac{2^p - 1}{2^r}}$ и назовем такие пиксели \textit{сдвиговыми}.

Покажем, что множество абсцисс сдвиговых пикселей, полученных согласно определению 2 равно $\hat{X}$.
Для того, чтобы пиксель с абсциссой $x$ был сдвиговым должно выполняться:
$$
\begin{cases} 
    \big[\frac{2^r x}{2^p - 1}\big] = m,          \\
    \big[\frac{2^r (x+1)}{2^p - 1}\big] = m + 1,  \\ 
    m \in [0, 2^r - 1],
\end{cases}
$$
где $m$ - целое число. Преобразуем данную систему:

$$
    \begin{cases} 
        \frac{2^r x}{2^p - 1} \geq m - \frac{1}{2},        \\
        \frac{2^r x}{2^p - 1} < m + \frac{1}{2},        \\
        \frac{2^r (x + 1)}{2^p - 1} \geq m + \frac{1}{2},  \\
        \frac{2^r (x + 1)}{2^p - 1} < m + \frac{3}{2},  \\
        m \in [0, 2^r - 1],
    \end{cases}
    \qquad
    \to
    \begin{cases} 
        x < (m + \frac{1}{2})\frac{(2^p - 1)}{2^r}      \\
        x > (m + \frac{1}{2})\frac{(2^p - 1)}{2^r} - 1  \\
        m \in [0, 2^r - 1].
    \end{cases}
$$
Множество решений системы можно записать в виде $X = \floor{(m + \frac{1}{2}) \frac{2^p - 1}{2^k}}$. 
Так как $X \equiv \hat{X}$ утверждение верно.
\end{proof}

Заметим, что любой дискретный паттерн, в частности диадический, с пикселем в начале координат можно задать множеством абсцисс сдвиговых пикселей.
Для получения ординаты любого пикселя паттерна необходимо проссумировать число сдвиговых пикселей, абсцисса которых меньше чем у данного. Назовем такую процедуру \textit{кумулятивное построение} дискретной прямой.

\begin{mydef}
    Дискретная прямая наклона $t \in [0, 2^p - 1]$, с двоичной записью $t = t_0t_1t_2...t_{p-1}$,
    
    \begin{equation}
        D(x,t) = \sum\limits_{r = 0}^{p - 1} t_r D_r(x)
    \end{equation}
    
    называется \textbf{диадическим паттерном}.
\end{mydef}

\begin{mystate}
    Для любого наклона $t \in [0, 2^p - 1]$ множества пикселей диадического паттерна полученные согласно определениям 1 и 3 соответственно.
\end{mystate}
\begin{proof}
    Согласно утверждению $1$, для случая базисных паттернов множество абсцисс сдвиговых пикселей задается формулой $X_{t} = \floor{(m + \frac{1}{2}) \frac{2^p - 1}{2^r}}$, где $m$ - номер сдвигового пикселя, $r$ - разряд ненулевого бита. 
    Множество сдвиговых пикселей для наклона произвольного $t$ есть объединение множеств сдвиговых пикселей для каждого базисного наклона $X = X_{t_0} \cup X_{t_1} \cup X_{t_2} \cup ... \cup X_{t_{p-1}}$.
    
    Кумулятивное построение по множеству $X$, очевидно, эквивалентно покоординатному суммированию соответствующих базисных диадических паттернов с соответствующими наклонами.
\end{proof}

Используя определение 3, запишем ошибку аппроксимации диадическим паттерном непрерывной прямой вида $L(y,t) = \frac{2^r x }{2^p - 1}$:

\begin{equation}
    E(x,t) = L(x,t) - D(x,t) = \sum\limits_{r = 0}^{p-1} t_r \bigg(\bigg[\frac{2^r x}{2^p - 1}\bigg] - \frac{2^r x}{2^p - 1}\bigg).
    \label{eq:error}
\end{equation}

Из формулы \ref{eq:error} видно, что наклон $t$ и абсцисса $x$ перестановочны:
\begin{equation*}
    E(x,t) = \sum\limits_{r = 0}^{p-1} t_r \bigg(\bigg[\frac{2^r x}{2^p - 1}\bigg] - \frac{2^r x}{2^p - 1}\bigg) = 
    \sum\limits_{r = 0}^{p-1} \sum\limits_{s = 0}^{p-1} t_r x_s \bigg(\bigg[\frac{2^r 2^s}{2^p - 1}\bigg] - \frac{2^r 2^s}{2^p - 1}\bigg).
    \label{eq:error_rev}
\end{equation*}

Другим существенным свойством диадического паттерна является симметрия ошибки аппроксимации, относительно центра паттерна.

\begin{mystate}
Для любого $y \in [0, 2^{p-1} - 1]$ и $t \in [0, 2^p - 1]$ верно, что $E(2^{p-1}-1-y,t) = - E(2^{p-1}+y,t)$.
\begin{proof}
        Перепишем утверждение согласно формуле \ref{eq:error}:
        \begin{equation*}
            \frac{(2^{p-1} - 1 - y)2^r}{2^p - 1} - \bigg[\frac{(2^{p-1} - 1 - y)2^r}{2^p - 1}\bigg] = 
            \bigg[\frac{(2^{p-1} + y)2^r}{2^p - 1}\bigg] - \frac{(2^{p-1} + y)2^r}{2^p - 1}.
        \end{equation*}
        Преобразуем данное выражение:
        \begin{equation*}
            \bigg[\frac{(2^{p-1} + y)2^r}{2^p - 1}\bigg] + \bigg[\frac{(2^{p-1} - 1 - y)2^r}{2^p - 1}\bigg] = 2^r.
        \end{equation*}
        Покажем, что данное уравнение есть тождество:
        \begin{equation*}
            \bigg[\frac{2^{p-1}2^r}{2^p-1} + \frac{2^ry}{2^p-1}\bigg] + 
            \bigg[\frac{2^{p-1}2^r}{2^p-1} - \frac{(y + 1)2^r}{2^p - 1}\bigg] = 2^r,
        \end{equation*}
        
        \begin{equation*}
            \bigg[2^{r-1} + \frac{2^{r-1}}{2^p-1} + \frac{2^ry}{2^p-1}\bigg] + 
            \bigg[2^{r-1} + \frac{2^{r-1}}{2^p-1} - \frac{(y + 1)2^r}{2^p - 1}\bigg] = 2^r.
        \end{equation*}
        В результате получим:
        \begin{equation*}
            \bigg[\frac{2^{r-1} + 2^ry}{2^p-1} \bigg] + 
            \bigg[\frac{-2^{r-1}-2^ry}{2^p-1}\bigg] = 0,
        \end{equation*}
        что есть тождественно верное уравнение,
        что и требовалось доказать. 
\end{proof}

\end{mystate}
\section{Вычислительные эксперименты: ошибка аппроксимации диадического паттерна}

В данной главе приводится ряд качественных и количественных результатов вычислительных экспериментов по исследованию структуры и точности аппроксимации диадичесого паттерна.

\begin{wrapfigure}{r}{0.5\textwidth}
    \centering
    \includegraphics[width=0.9\linewidth]{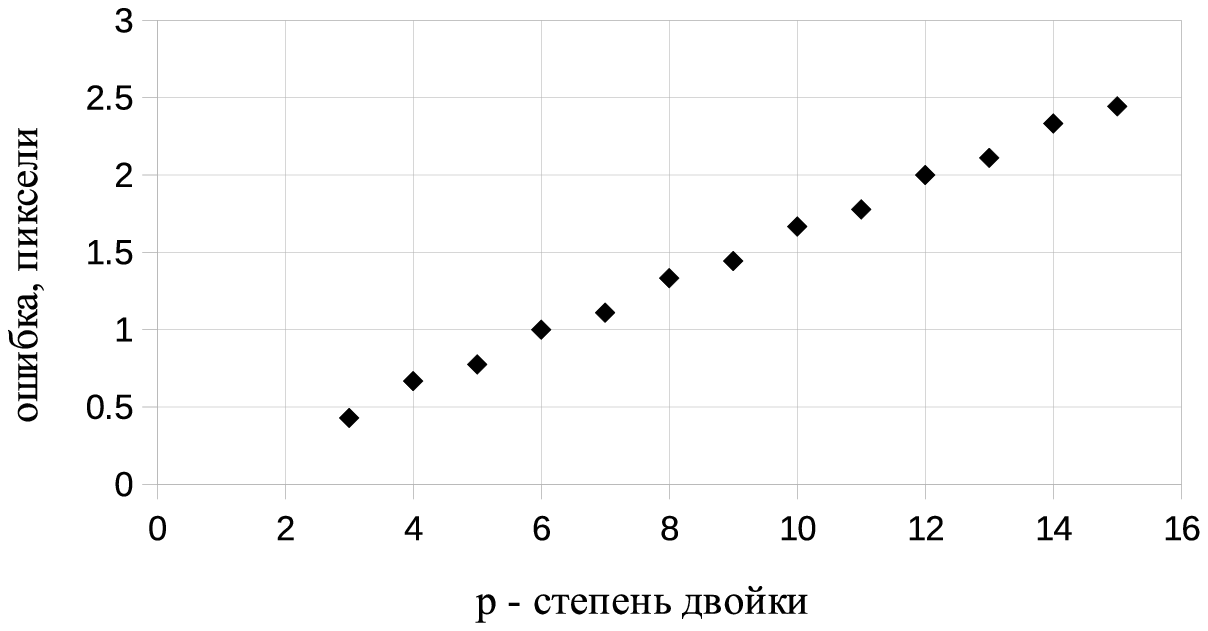}
    \caption{Зависимость максимальной ошибки аппроксимации диадическим паттерном от логарифма размера изображения.}
    \label{fig:error_p}
\end{wrapfigure}

Ошибка аппроксимации определена формулой \ref{eq:error}.
Из формулы \ref{eq:error} видно, что верхняя граница оценки точности аппроксимации составляет $\frac{p}{2}$, так как модуль каждого слагаемого не превосходит $\frac{1}{2}$.
Однако полученные вычислительные результаты \cite{Gotz1995, Brady1998} заставляют предположить, что оценка $\frac{p}{2}$ слишком груба (смотри рисунок \ref{fig:error_p}). 
Дальнейшее изложение работы посвящено доказательству того, что для изображений со стороной $2^{2m}$ пикселей точная оценка ошибки аппроксимации - $\frac{p}{6}$, для изображений со стороной $2^{2m + 1}$ - $\frac{p}{6} - \frac{2^p+1}{18(2^p-1)}$ \cite{ershov2015dev}, где $m \in [0, \frac{p}{2}]$.

\begin{figure}[h!]
    \centering
    \includegraphics[scale=0.3]{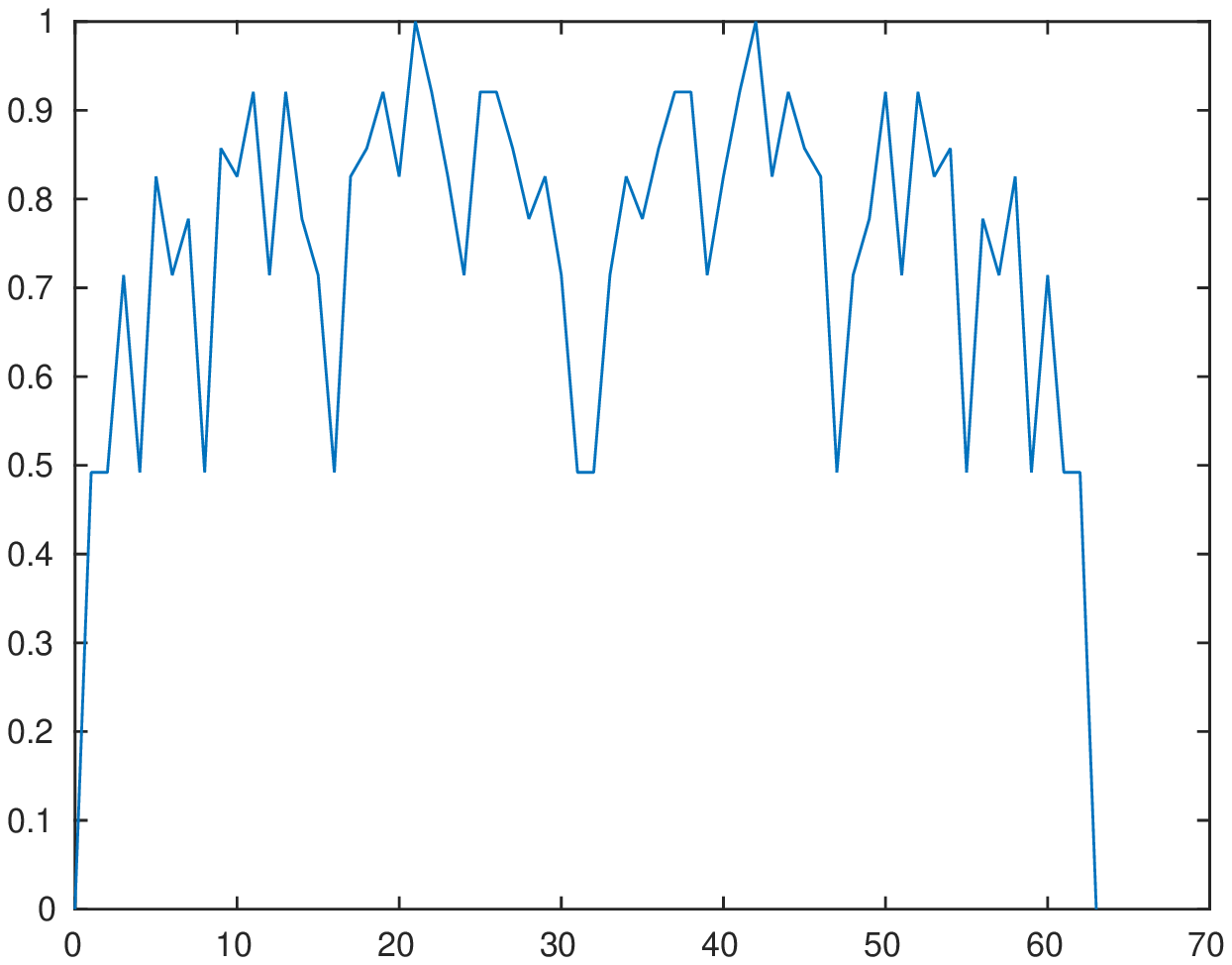}
    \includegraphics[scale=0.3]{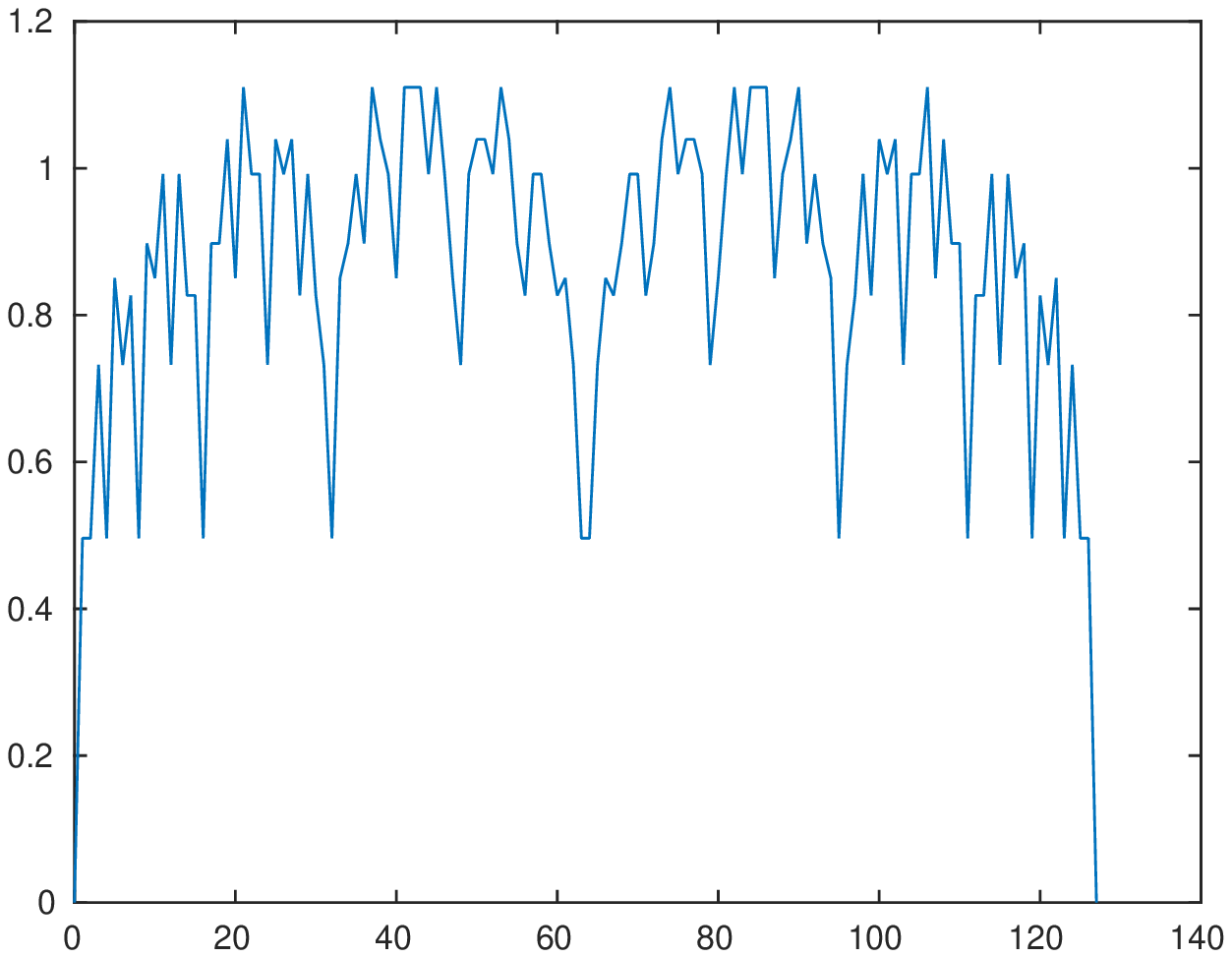}
    \includegraphics[scale=0.3]{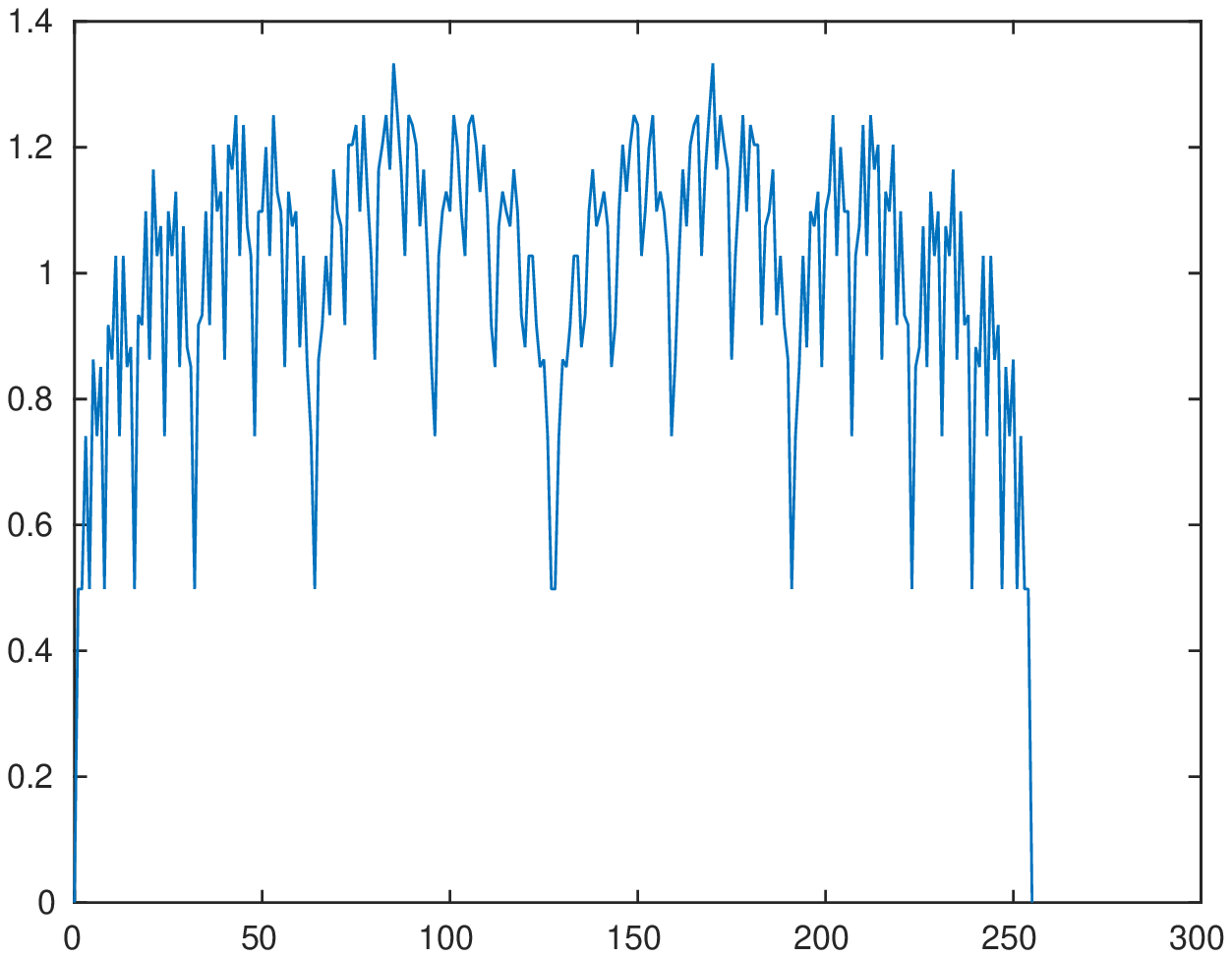}
    \caption{Зависимость максимальной ошибки аппроксимации для каждого наклона для размеров изображений $2^6$, $2^7$, $2^8$ соответственно: по горизонтальной оси - наклон, по вертикальной - величина ошибки.}
    \label{fig:whall}
\end{figure}

Кроме того, производилось исследование величины максимальной ошибки для каждого наклона диадического паттерна. 
Данная ошибка имеет фрактальную природу, при том для чётных $p$ ошибка достигает пика только в двух точках, в то время как для нечетных $p$ - имеется $2(p - 1)$ пиков (смотри рисунок \ref{fig:whall}.). 

Получено распределение ошибки аппроксимации диадическими паттернами всех наклонов $t$. 

\begin{figure}[h!]
    \centering
    \includegraphics[scale=0.3]{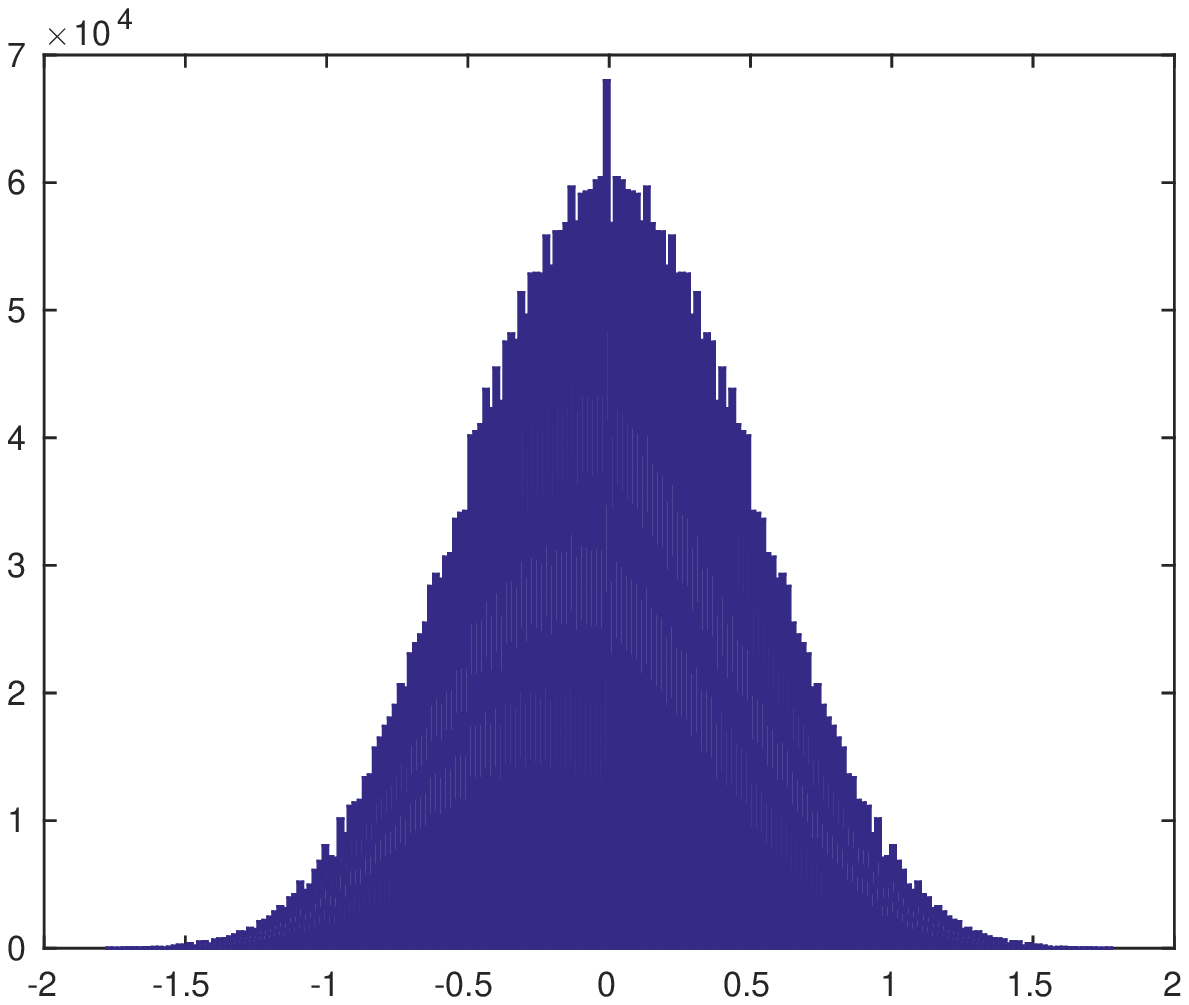}
    \includegraphics[scale=0.3]{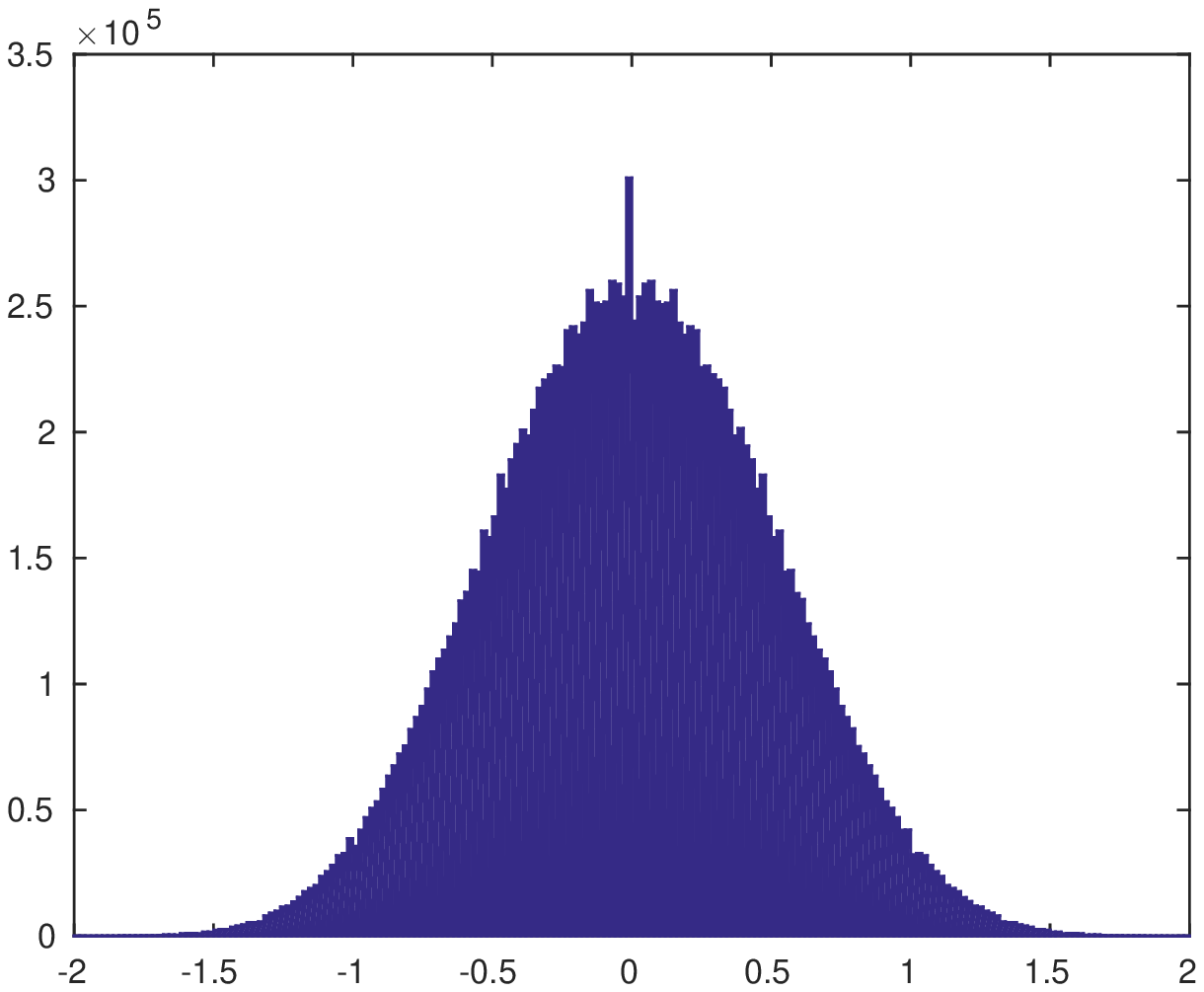}
    \includegraphics[scale=0.3]{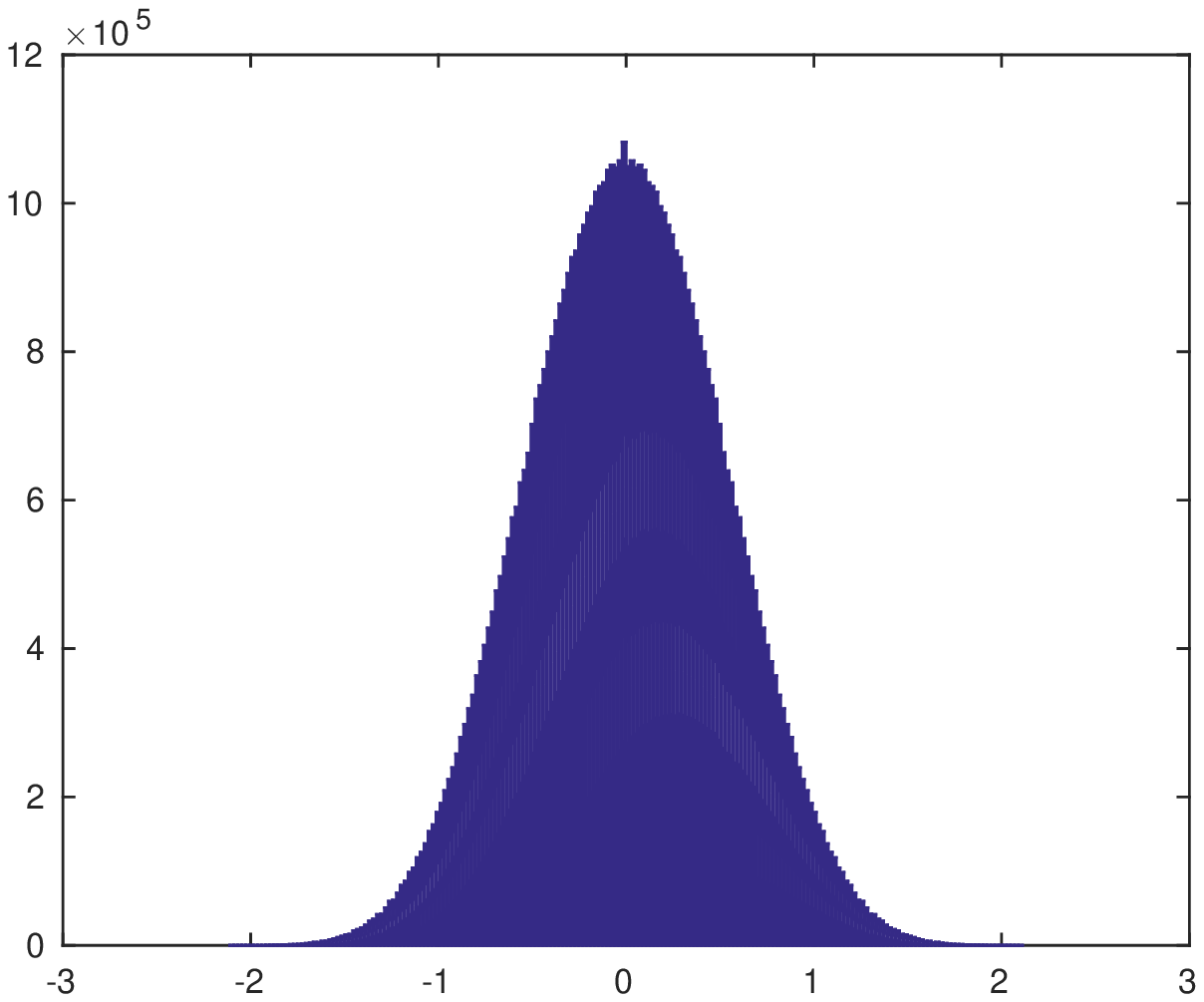}
    \caption{Распределение ошибки аппроксимации диадических паттернов всех наклонов для изображения с линейным размером $2^{11}, 2^{12}, 2^{13}$ соответственно.}
    \label{fig:gauss}
\end{figure}

\section{Максимальная ошибка аппроксиации $p/6$}

Запишем выражение ошибки аппроксимации диадическим паттерном, согласно выражению \ref{eq:error}:
\begin{equation}
    E(x,t) = \sum\limits_{r = 0}^{p - 1} t_r E_r(x),
    \label{eq:error_simple}
\end{equation}
где $E_r(x) = \bigg(\bigg[\frac{2^r x}{2^p - 1}\bigg] - \frac{2^r x}{2^p - 1}\bigg)$.

Заметим, что для фиксированного $x$ модуль ошибки $E(x,t)$ будет максимален, когда $t_r \neq 0$ только для слагаемых одного знака.
Так как, согласно утверждению 2.3 ошибка аппроксимации симметрична, далее будем рассматривать только такие $(r,x)$ для которых $E_r(x) < 0$.
Иными словами достаточно выбрать те $r$, при которых \textit{двоичное представление числа} $s_r = 2^r x$ длины $p$ имеет ноль в старшем разряде.
Двоичное представление $s_{r+1}$ может быть получено из $s_r$ циклическим сдвигом на один разряд.

Поэтому сформируем таблицу $T(x)$ всевозможных циклических сдвигов двоичной записи $x$:
\begin{center}
    \begin{tabular}{ c c c c }
        $x_{p-1}$ & $x_{p-2}$ & ... & $x_0$       \\
        $x_{p-2}$ & $x_{p-3}$ & ... & $x_{p-1}$   \\
        $x_{p-3}$ & $x_{p-4}$ & ... & $x_{p-2}$   \\
        ...       & ...       & ... & ...         \\
        $x_0$     & $x_{p-1}$ & ... & $x_{1}$
    \end{tabular}
\end{center}

Определим функцию от таблицы $S(T(x))$, как арифметическую сумму по строкам с нулевым старшим битом.
Итак, поиск максимума модуля ошибки эквивалентен решению задачи:

\begin{equation}
S(T(x)) \to \max\limits_x
\end{equation}

\begin{mydef}
Максимайзером будем называеть любой $\hat{x}$ такой, что

    \begin{equation}
        \hat{x} = \argmax\limits_{x \in [0, 2^p - 1]} S(T(x)).
        \label{eq:x}
    \end{equation}

\end{mydef}

Проанализируем структуру таблицы $T(x)$.
Символами \lal и \rands обозначим произвольную двоичную последовательность. 
Символами \revlal и \revrands обозначим инверсию, соответственно, последовательностей \lal и \rands, записанных в обратном порядке. 
Например, если \lal = $010000$, \revlal = $111101$.

По построению, для любого числа $x$ таблица $T(x)$ будет иметь следующую структуру, где \lal = $x_{p-2},...,x_p$:
\begin{center} 
    \begin{tabular}{ c c c c }
        $x_{p-1}$                & \multicolumn{3}{c}{\lal}       \\
        \multirow{4}{*}{\rotlal} & $x_{p-3}$ & ... & $x_{p-1}$   \\
                                 & $x_{p-4}$ & ... & $x_{p-2}$   \\
                                 & ...       & ... & ...         \\
                                 & $x_{p-1}$ & ... & $x_{1}$,
    \end{tabular}
    \label{tab:initial}
\end{center}
заметим, что битовое представление $x$ содержится как в первой строке, так и в первом столбце.

\begin{mylemm}
    О замене.

    Пусть символ $T_\epsilon(x)$ обозначает таблицу вида с $x_{p-1} = \epsilon$:

    \begin{center}
        \begin{tabular}{ c c c c }
            $\epsilon$               & \multicolumn{3}{c}{\lal}      \\
            \multirow{4}{*}{\rotlal} & $x_{p-3}$ & ... & $\epsilon$   \\
                                     & $x_{p-4}$ & ... & $x_{p-2}$   \\
                                     & ...       & ... & ...         \\
                                     & $\epsilon$ & ... & $x_{1}$,
        \end{tabular}
    \end{center}
    тогда $S(T_0 (x)) > S(T_1 (x))$ эквивалентно \lal > \revlal.
\end{mylemm}
\begin{proof}
    Для доказательства леммы достаточно показать, что 
    \begin{center}
        $S(T_0 (x)) - S(T_1 (x)) =$ \lal $-$ \revlal. 
    \end{center}   
    Так как суммация производится только по строкам с нулевым старшим битом, то разность первых строк равна \lal.
    Далее, для пары строк с номером $r > 1$ в таблице $T_0(x)$ значение в разряде $p-r$ равно нулю, в то время как в правой таблице $T_1(x)$ -- единице, значения в остальных разрядах строк совпадают, т.е. вклад такой пары строк есть отрицательное число с единственной единицей в $p-r$ разряде. 
    Тогда сумма вкладов всех таких пар строк есть $-$\revlal.   
    Что и требовалось доказать.
\end{proof}

\begin{mylemm}
    О перестановке.

    Пусть символ $1_p$ обозначает битовую последовательность единиц длинны $p$, а символ $T_{\epsilon\delta}(x)$ - таблицу вида:

    \begin{center}
        \begin{tabular}{ c c c c }
            $\epsilon$               & $\delta$  & \multicolumn{2}{c}{\lal} \\
            $\delta$                 & $x_{p-3}$ & ... & $\epsilon$          \\
            \multirow{3}{*}{\rotlal} & $x_{p-4}$ & ... & $\delta$          \\
                                     & ...       & ... & ...                \\
                                     & $\epsilon$ & ... & $x_{1}$,
        \end{tabular}
    \end{center}
    тогда $S(T_{01} (x)) > S(T_{10} (x))$ эквивалентно \lal + \revlal < $1_p$.
\end{mylemm}
\begin{proof}
    Аналогично доказательству леммы 1, запишем разность сумм по таблицам:
    
    \begin{center}
        $S(T_{01} (x)) - S(T_{10} (x)) = $ 01\lal $-$ 0\lal1 $-$ \revlal,
    \end{center}
    здесь первое слагаемое есть вклад пары строк с $r = 0$, второе слагаемое -- пары строк с $r = 1$, а третье слагаемое -- это вклад оставшихся пар строк, полученный аналогичным лемме 1 способом. После преобразования получим:

    \begin{center}
        $S(T_{01} (x)) - S(T_{10} (x)) = 1_p -$ \lal $-$ \revlal,
    \end{center}
    что и требовалось.
\end{proof}

Далее последовательно проверим всевозможные $x$ на предмет принадлежности к множеству максимайзеров.
Обозначим символом \vave{m} последовательность чередующихся нулей и единиц длинны $m$, а символом \vavep{m}{\epsilon}{\delta} последовательность чередующихся нулей и единиц длинны $m$ со значениями $\epsilon$, $\delta$ в старшем и младшем разрядах соответственно.
Напирмер \vavep{5}{1}{1} $= 10101$.


\begin{mystatet}
    Если в двоичной записи числа $x$ встречаются три подряд идущих равных бита (далее \textit{битовая тройка}), то такой $x$ не может быть максимайзером.
\end{mystatet}
\begin{proof}
    Без ограничения общности рассмотрим число в котором есть как минимум одна битовая тройка единиц $111\rands$. 
    Пусть $\hat{x} = 11\rands1$. Так как $T(x)$ содержит все циклические сдвиги, то $S(T(x)) = S(T(\hat{x}))$, следовательно, достаточно показать, что 
    \begin{center}
        $S(T(11$\rands$1))$ < $S(T(01$\rands$1))$.
    \end{center}
    Обозначим \lal = $1$\rands$1$, тогда \revlal = $0$\revrands$0$, видно, что \lal > \revlal.
    Следовательно, согласно лемме 1, $S(T(11$\rands$1)) < S(T(01$\rands1$))$ и $111\rands$ не может быть максимайзером.
\end{proof}


\begin{mystatet}
    Число вида $x = $ 11 \vavep{l}{0}{0}11\vavep{m}{0}{0}11\rands для любых положительных $l,m$ не может быть максимайзером.
\end{mystatet}
\begin{proof}

Пусть $\hat{x}$ получен в результате циклического сдвига $x$:
\begin{center}
    $\hat{x} = $ 11 \vavep{m}{0}{0}11\rands11\vavep{l}{0}{0}.
\end{center}

Обозначим
\begin{center}
    \lal = 1 \vavep{m}{0}{0}11\rands11\vavep{l}{0}{0} = \vavep{m+2}{1}{1}1\rands11\vavep{l}{0}{0},
\end{center}
тогда
\begin{center}
    \revlal = \vavep{l}{1}{1}00\revrands00\vavep{m}{1}{1}0,
\end{center}

Для сравнения \lal и \revlal необходимо рассмотреть три случая: $m + 2 < l$, $m + 2 = l$, $m + 2 > l$.
Рассмотрим, например, случай, когда $m + 2 < l$. 
Будем побитово сравнивать \lal и \revlal начиная со старшего бита.
Видно, что в первом несовпавшем разряде \lal содержит единицу, а \revlal - ноль, следовательно, \lal > \revlal. 
Аналогичным образом рассматриваются остальные два случая.
Следовательно, согласно лемме 1:
    \begin{center}
        $S(T(11$ \vavep{l}{0}{0}11\vavep{m}{0}{0}11\rands$))$ < 
        $S(T(01$ \vavep{l}{0}{0}01\vavep{m}{0}{0}11\rands$))$,
    \end{center}
то есть $x$ не является максимайзером.
\end{proof}


\begin{mystatet}
    Если в битовом представлении $x$ содержится подпоследовательность вида $1100$, то $x$ не может быть максимайзером.
\end{mystatet}
\begin{proof}
    Пусть $x = 100\rands1$, а $\hat{x} = 010\rands1$.
    
    Пусть \lal = 0\rands1, тогда \revlal = 0\revrands1. 
    Очевидно, что \lal + \revlal < $1_{p-2}$, так как у обоих слагаемых старший разряд имеет нулевое значение.
    
    Тогда, согласно лемме 2, $S(T(x)) < S(T(\hat{x}))$ и, следовательно, $x$ не может быть максимайзером.
\end{proof}


\begin{mystatet}
    Число вида $x = 11$\vavep{m}{0}{0}11\vavep{n}{0}{1}00\rands не может быть максимайзером.
\end{mystatet}
\begin{proof}
    Пусть $\hat{x}$ - циклический сдвиг числа $x$:
    \begin{center}
        $\hat{x} = 10$\vavep{n-1}{1}{1}$00\rands11$\vavep{m}{0}{0}$1$.
    \end{center}
    Докажем данное утверждение для $\hat{x}$, заметим, что $x = 10$\lal, где
    \begin{center}
        \lal = $1$\vavep{n-2}{0}{1}$00\rands11$\vavep{m}{0}{0}$1$,
    \end{center}
    тогда 
    \begin{center}
        \revlal = 0\vavep{m}{1}{1}$00$\revrands$11$\vavep{n-2}{0}{1}0.
    \end{center}
    Видно, что независимо от соотношения величин $n - 2$ и $m$, \lal + \revlal < $1_{p-2}$.
    Следовательно, согласно лемме 2, $S(T(10$\vavep{n-1}{1}{1}$00\rands11$\vavep{m}{0}{0}$1$)) < $S(T(00$\vavep{n-1}{1}{1}$00\rands11$\vavep{m}{0}{0}$1$)) и $\hat{x}$ - не максимайзер.
\end{proof}


\begin{mystatet}
    Число вида $x = 11$\vavep{m+k}{0}{1}$00$\vavep{m}{1}{0}11\rands не может быть максимайзером ($m>1$).
\end{mystatet}
\begin{proof}
    Пусть $\hat{x}$ - циклический сдвиг числа $x$:
    \begin{center}
        $\hat{x} = 01$\vavep{m-1}{0}{0}$11\rands11$\vavep{m+k}{0}{1}$0$,
    \end{center}
    докажем данное утверждение для $\hat{x}$, так как $S(T(x)) = S(T(\hat{x}))$. 
    Пусть
    \begin{center}
        \lal = \vavep{m-1}{0}{0}$11\rands11$\vavep{m+k}{0}{1}$0$,
    \end{center}
    тогда 
    \begin{center}
        \revlal = \vavep{m+k+1}{1}{1}$00$\revrands$00$\vavep{m-1}{1}{1}.
    \end{center}
    
    Поскольку для любых положительных $m,n$ верно $m + k + 1 > m - 1$, ясно, что \lal + \revlal > $1_p$. 
    Следовательно, согласно лемме 2, $x$ не может быть максимайзером.
\end{proof}


\begin{mytheo}
    Максимальная ошибка аппроксимации диадическим паттерном достигается при $x = \big[{\frac{2^p}{3}}\big]$ и $x = \big[{\frac{2^{p+1}}{3}}\big]$ при наклонах $t = \big[{\frac{2^p}{3}}\big]$ и $t = \big[{\frac{2^{p+1}}{3}}\big]$, и её модуль равен $\frac{p}{6}$, для чётных $p$.
\end{mytheo}
\begin{proof}
    Ясно, что $x_1 = $ \vavep{p}{0}{1} $ = \big[{\frac{2^p}{3}}\big]$,  $x_2 = $ \vavep{p}{0}{1} $ = \big[{\frac{2^{p+1}}{3}}\big]$.
    Любые другие $x$ не могут быть максимайзерами, согласно утверждениям $1-5$.
    Заметим, что $x = 11$\vave{n}$11$\vave{m} является частным случай утверждения 2, а $x = 11$\vave{n}$00$\vave{m} являются частным случем утверждения 4.
    
    В силу симметрии (см утверждение 2.3) $t_1 = \big[{\frac{2^p}{3}}\big]$, $t_2 = \big[{\frac{2^{p+1}}{3}}\big]$ являются наклонами для данных максимайзеров, а для каждого из этих наклонов пиковая ошибка достигается при $x_1 = \big[{\frac{2^p}{3}}\big]$, $x_2 = \big[{\frac{2^{p+1}}{3}}\big]$.
    
    Рассчитаем теперь значение пиковой ошибки, к примеру для случая $x_1 = \big[{\frac{2^p}{3}}\big]$.
    Для этого достаточно вычислить сумму по таблице $S(T(x_1))$ и поделить результат на $2^p - 1$, согласно формуле \ref{eq:error}.
    Сумма по одной значимой строке в таблице, есть сумма геометрической прогрессии из $\frac{p}{2}$ элементов со знаменателем $q = 4$, с первым членом, равным $1$. 
    Нетрудно вычислить, что сумма по одной строке равна $s = \frac{4^{p/2} - 1}{3} = \frac{2^p - 1}{3}$, тогда для всех $\frac{p}{2}$ значимых сумма по таблице равна $\frac{p(2^p - 1)}{6}$.
    В результате пиковая ошибка аппроксимации, согаслно формуле \ref{eq:error_simple} равна $\frac{p}{6}$.
    
    Аналогично доказательство применимо для остальных пар $(x,t)$, где достигается пиковое значение ошибки.
\end{proof}

\begin{mytheo}
    Модуль максимальной ошибка аппроксимации диадическим паттерном равен $\frac{p}{6} - \frac{1}{18}$, для нечётных $p \to \infty$ и достигается при $x = 00$\vavep{p-2}{1}{1} или $x = 11$\vavep{p-2}{0}{0}.
\end{mytheo}
\begin{proof}
    Согласно утверждениям 1-5 для случая нечетных $p$ максимайзером может быть только число, в битовом представлении которого содержится только одна пара соседствующих единичных или нулевых битов.
    
    Для определения величины этой ошибки при $p \to \infty$ аналогично Теореме 1 подсчитаем сумму по таблице. 
    Пусть $x_1 = $\vavep{p-2}{0}{0}$11$ и $t_1 = $\vavep{p-2}{0}{0}$11$.
    Видно, что такое число содержит $m = \frac{p-1}{2}$ нулей и $n = \frac{p-1}{2} + 1$ единиц. 
    
    Для начала вычислим сумму $S_1$ по первой строке таблицы. 
    Ясно, что \vavep{p-2}{0}{0}$11$ = \vavep{p}{0}{0} + 1. Заметим, что сумма по \vavep{p}{0}{0} является суммой геометрической прогрессии с перым элементом равным $a_1 = 2$ и знаменателем $q = 4$, и равна $\frac{2}{3} (2^{p-1} - 1)$. Тогда $S^1 = \frac{2}{3} (2^{p-1} - 1) + 1$.
    
    Заметим, что разность соседних значимых строк $\Delta_i$ в таблице формирует геометрическую последовательность, $\Delta_1 = T_1(x_1) - T_3(x_1) = 10$, $\Delta_2 = T_3(x_1) - T_5(x_1) = 1000$, обобщая, $\Delta_i = 2\cdot4^{i-1}$, где $i \in [1, \frac{p-1}{2}-1]$.
    
    Тогда, значение суммы по каждой значимой строке $k$ (индекс пробегает по строкам с нулевым старшим битом) можно представить как сумму по первой строке и добавки к каждой строке, равной соответствующей сумме $S_k = \sum\limits_{i = 1}^{k-1}\Delta_i$. 
    Таким образом сумма по таблице равна 
    
    \begin{equation*}
        S = \bigg[\frac{p-1}{2} \cdot S^1 + \sum\limits_{k = 1}^{p - 1} S_k\bigg].
    \end{equation*}
    Преобразовав данное выражение, получим 
    
    \begin{equation}
        S = \bigg[\frac{p-1}{6} (2^p - 1) + \frac{p-1}{3} + \frac{1}{9} (2^p - 1) - \frac{7}{9} - \frac{2}{3} (p - 1) \bigg]. 
        \label{eq:10}
    \end{equation}
    
    Тогда, в результате деления выражения \ref{eq:10} на $2^p - 1$, согласно формуле \ref{eq:error_simple}, и устремления $p \to \infty$ получим $E(x_1,x_2) = \frac{p}{6} - \frac{1}{18}$, что и требовалось доказать.
    
    Аналогичным образом производится доказательство и для всех остальных пар $(x, t)$, где достигается пиковое значение ошибки.
\end{proof}

\section{Связь с одномерной моделью Изинга для антиферромагнетка}

Одним из интересных свойств предложенного в статье функционала по таблице является его эквивалентность функционалу в модели Изинга \cite{pirogovizing} для одномерного кольцевого антиферромагнетика в отсутствии внешнего поля с дальним взаимодействием.

Подсчитаем арифметическую сумму всех циклических сдвигов $x$, начинающихся на $0$:
\begin{equation*}
    S(T(x)) = F(x) = \sum\limits_i \sum\limits_j (1 - \epsilon_i)\epsilon_j \frac{1}{2^{d(i,j)}},
\end{equation*}
где $d(i,j)$ - циклическое вправо расстояние.

Сделаем замену переменных 
\begin{equation*}
    \xi_i = 2\epsilon_i - 1.
\end{equation*}

Тогда
\begin{equation*}
    \begin{split}
        F(x) = \sum\limits_i \sum\limits_j \frac{1 - \xi_i}{2}\frac{\xi_j + 1}{2} \frac{1}{2^{d(i,j)}} = \\
        =\frac{1}{4} \sum\limits_i \sum\limits_j (1 + (\xi_j - \xi_i) - \xi_i \xi_j) \frac{1}{2^{d(i,j)}} = \\
        = const - \sum\limits_i \sum\limits_j \frac{\xi_i\xi_j}{2^{d(i,j)}}.    
    \end{split}
\end{equation*}

Функционал $F(x) = F(\epsilon_0, ..., \epsilon_{p-1}) = const - E(\xi_0, ..., \xi_{p-1})$, таким образом максимизация $F$ на $\epsilon \in \{0, 1 \}^p$ эквивалентна минимизации $E$ на $\xi \in \{-1, 1\}^p$:
\begin{equation*}
    E(S) = \sum\limits_i \sum\limits_j \frac{\xi_i \xi_j}{2^{d(i,j)}}.
\end{equation*}

Каждая пара взаимодействует дважды, как $(i,j)$ и как $(j,i)$ так что справедлива следующая формула:
\begin{equation*}
    E(S) = - \sum\limits_{i < j} \xi_i \xi_j \bigg(\frac{1}{2^{d(i,j)}} + \frac{1}{2^{d(j,i)}}\bigg).
\end{equation*}

Итак, задача поиска пиковой ошибки эквивалентна задаче минимизации энергии антиферромагнетика с дальним взаимодействием и потенциалом $P_{ij} = \frac{1}{2^{d(i,j)}} + \frac{1}{2^{d(j,i)}}$.
Расстояния вычисляются следующим образом
\begin{equation*}
    \{d(i,j), d(j,i)\} = \{a = max(i,j) - min(i,j), p - a \},
\end{equation*}
где $p = \log_2 n$, $n$ - линейный размер изображения.
\section{Заключение}

В работе проведены вычислительные и аналитические исследования структуры диадического паттерна, в частности ошибка его аппроксимации геометрической прямой. 
Показано, что пиковая ошибка для изображения линейным размером $n$ не превосходит $\frac{\log(n)}{6}$ для случая четных $n$, и $\frac{\log(n)}{6} - \frac{1}{18}$ для нечетных. 
Найдена аналогия между предложенным функционалом по таблице и одномерной моделью Изинга.

\bibliography{main}{}
\bibliographystyle{plain}

\end{document}